\documentclass[runningheads]{llncs}
\usepackage[T1]{fontenc}
\usepackage{graphicx}

\begin{document}
\title{CARJAN: Agent-Based Generation and  
       Simulation of Traffic Scenarios with AJAN}
\titlerunning{CARJAN}
%

\author{Leonard Frank Neis 
        \and Andr\'e Antakli 
        \and Matthias Klusch}

\authorrunning{L. Neis et al.}
%
\institute{German Research Center for Artificial Intelligence (DFKI) \\
           Saarland Informatics Campus, Saarbruecken, Germany\\
          \{leonard\_frank.neis, andre.antakli, matthias.klusch\}@dfki.de\\
          }
\maketitle              
\begin{abstract}
User-friendly modeling and virtual simulation of urban traffic scenarios
with different types of interacting agents such as pedestrians, cyclists and 
autonomous vehicles remains a challenge. We present CARJAN, a novel tool for 
semi-automated generation and simulation of such scenarios based on the 
multi-agent engineering framework AJAN and the driving simulator CARLA. 
CARJAN provides a visual user interface for the modeling, storage and 
maintenance of traffic scenario layouts, and leverages SPARQL Behavior Tree-based 
decision-making and interactions for agents in dynamic scenario simulations in CARLA. 
CARJAN provides a first integrated approach for interactive, intelligent agent-based 
generation and simulation of virtual traffic scenarios in CARLA. 
\end{abstract}
%
%
\section{Introduction}

User-friendly modeling and virtual simulation of urban traffic scenarios
with different types of interacting agents such as pedestrians, cyclists and 
autonomous vehicles \cite{EuroPVI21} remains a challenge. 
There are a few frameworks available for the generation and simulation of 
virtual traffic scenarios in the driving simulator CARLA, such as SCENIC 
\cite{SCENIC23,SCENIC19} and OASIS \cite{Carla-OASIS,OASIS25}.
However, these approaches lack the means of a more explainable, 
declarative behavior modeling of intelligent agents and their dynamic interactions, 
such as pedestrian-vehicle interactions in urban traffic scenarios.

\noindent
To this end, we developed CARJAN, a tool for visual, semi-automated modeling and 
simulation of interactive agent-based traffic scenarios in CARLA \cite{thesisneis}. 
CARJAN builds on the multi-agent framework AJAN \cite{AJAN} to enable a declarative behavior 
modeling of agents with SPARQL-extendet Behavior Trees for an event-driven, 
semantic context-aware decision-making in dynamic scenarios. 
CARJAN integrates agent-based scenario generation and live simulation in CARLA 
in one workflow and graphical user interface.  

\noindent
The remainder of this paper is structured as follows.
In Section 2, we recall relevant basics of the used multi-agent system engineering 
framework AJAN and the driving simulator CARLA, and describe the functionalities 
of our CARJAN tool for integrated scenario generation and simulation as an extension 
of AJAN connected with CARLA in Section 3. We then briefly differentiate CARJAN 
from existing relevant CARLA scene generation frameworks in Section 4, 
and conclude in Section 5.

\section{Background}

\subsection{AJAN Agent Engineering Framework}

\noindent
AJAN, short for Accessible Java Agent Nucleus, offers a resilient agent-based framework for agent 
engineering \cite{AJAN}. It supports semantic reasoning over RDF data and enables the execution of agent behavior modeled through SPARQL-extendet Behavior Trees. The platform integrates a triple store for managing knowledge graphs and provides a runtime environment for agents to execute actions, query knowledge, and respond to changes in their environment. AJAN exposes its components via RESTful APIs, allowing external systems to interact with agents. CARJAN leverages the AJAN core framework as its foundation. 

\noindent
\textbf{AJAN Agent Behavior:} Behavior Trees (BTs) are a modular and hierarchical model for describing agent behavior, widely used in robotics and game AI \cite{Marzinotto+14}. In CARJAN, agent behaviors are defined using SPARQL queries structured according to the BT formalism. Each node in the tree corresponds to a SPARQL construct that can evaluate conditions, trigger actions, or control the flow of execution. This declarative approach allows for flexible, modular behavior modeling, where decision-making is directly tied to the agent’s knowledge graph. The use of SPARQL enables real-time reasoning and allows agents to adapt their behavior based on dynamically updated semantic information.

\vspace{-0.5cm}
\begin{figure}
    \centering
    \includegraphics[width=1\linewidth]{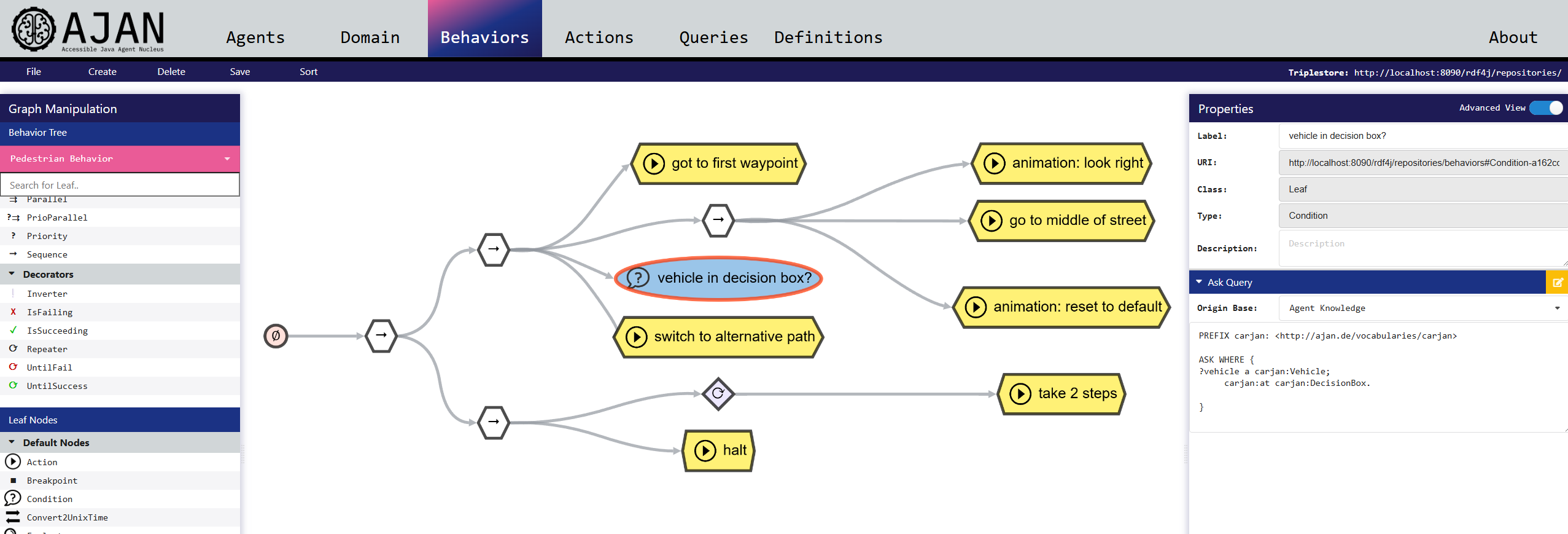}
    \vspace{-0.5cm}
    \caption{Agent behavior view of the AJAN-Editor with modeled SPARQL Behavior Tree of a pedestrian for crossing the street while checking for approaching cars.}
    \label{fig:pedestrian-BT}
\end{figure}
\vspace{-0.5cm}

\noindent
\textbf{AJAN-Editor:} The AJAN-Editor is an open-source, web-based environment for modeling, executing, and debugging agents in the AJAN framework. Implemented as a browser-based web service, it interfaces with the AJAN-Service and RDF triplestores to manage RDF-defined agent models—endpoints, events, goals, and SPARQL Behavior Trees—with live monitoring and execution tracing. In addition, agent knowledge queries and manipulation can be defined, and interaction with the agent environment can be specified. The core feature is the graphically based behavior modeling of SPARQL Behavior Trees using drag'n'drop. Figure \ref{fig:pedestrian-BT} shows a SPARQL-driven SPARQL Behavior Trees in the AJAN Editor controlling a simulated pedestrian. Yellow action nodes make the pedestrian cross while looking over their shoulder; if a blue condition detects an approaching car, the agent returns to the original sidewalk to avoid an accident.

\subsection{CARLA Driving Simulator}

\noindent
CARLA is an open-source simulation platform widely used for research on autonomous driving, 
especially in urban environments \cite{CARLA}. It follows a modular client-server architecture 
where the server handles rendering, physics, and dynamic interactions, while a Python API on the 
client-side enables scenario creation and runtime control of agents and environmental conditions. 
In its first version, CARJAN relies on CARLA version 0.9.15 built on Unreal Engine 4, 
which provides realistic urban environments and a variety of entities to model traffic situations 
in a game-like engine. In addition, CARLA offers control endpoints and sensory data such as LiDAR 
simulation in real-time. 

\noindent
However, scenario modeling via CARLA's native Python endpoint is typically static, code-intensive and 
error-prone, where even minor scripting errors can cause abrupt simulator terminations without clear 
diagnostics. Moreover, the manual scripting required for defining agent behaviors and environmental 
settings makes scenario creation cumbersome, particularly for nuanced pedestrian-vehicle scenarios. 
These drawbacks stress the necessity for more structured and visual approaches to scenario modeling.

\section{CARJAN}

\subsection{Architecture}

\noindent
The tool CARJAN combines services of the multi-agent system engineering framework AJAN 
and the driving simulator CARLA connected via Flask\footnote{Flask:
\url{https://flask.palletsprojects.com/en/stable/}} to enable an integrated 
generation and simulation of traffic scenarios in one graphical user interface (cf. Fig. 1). Scenarios in CARJAN are visually modeled by the user with grid-based layouts 
and placing dynamic and static entities in them. 
Intelligent agents and their behaviors are designed with Behavior Trees (BTs)\cite{BT18,BT20} 
and functional plug-ins in AJAN. 
CARJAN represents the modeled scenarios internally in RDF, and leverages 
via its middleware component {\tt carjanService} the Flask framework (a) to convert them 
into CARLA scenarios, and (b) to execute AJAN agent commands in CARLA. 
The first version of CARJAN is open-source available\footnote{CARJAN v1: 
\url{https://github.com/leonardneis/CARJAN}}. 

\begin{figure}[hbtp!]
\includegraphics[width=\textwidth]{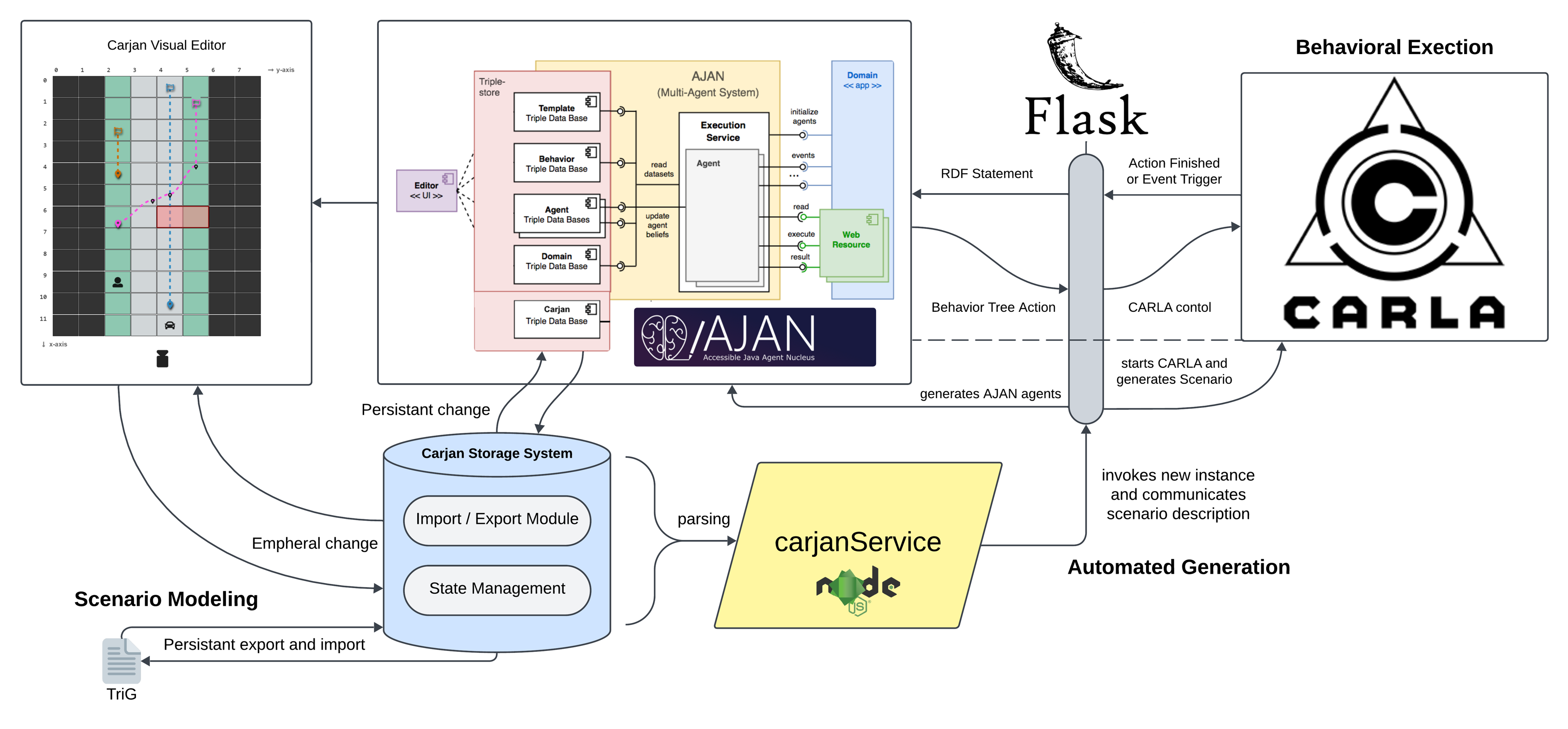}
\caption{CARJAN architecture overview: CARJAN embedded into AJAN Editor for 
intgegrated modeling of scenarios and agents, and connected via Flask with driving 
simulator CARLA.} 
\label{fig1}
\end{figure}
\noindent

\subsection{Scenario Generation}
%
\noindent
CARJAN provides the user with a GUI for the visual modeling and maintaining of traffic 
scenarios with agents such as pedestrians and vehicles (cf. Fig. 2). The CARJAN middleware 
service {\tt carjanService} seamlessly interfaces between this visual scenario editor, 
the AJAN agent execution backend and the CARLA simulation environment via dedicated Flask 
services to allow for an automated translation to and GUI-integrated scenario testing in CARLA. 
%
\begin{figure}[hbtp!]
\includegraphics[width=\textwidth]{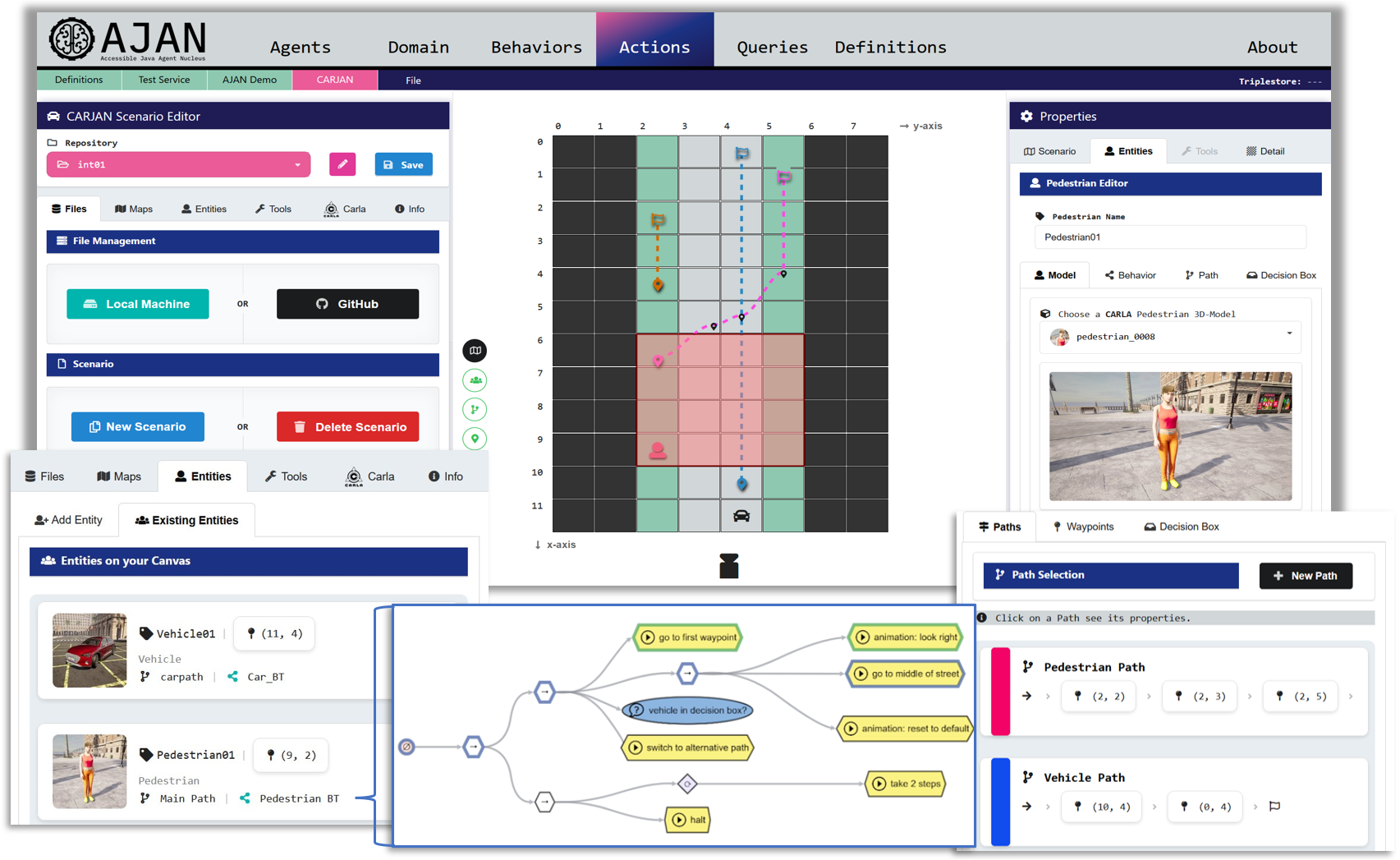}
\caption{CARJAN GUI for modeling and simulation of traffic scenarios and agents.} 
\label{fig2}
\end{figure}
%

\subsubsection{Modeling of traffic scenarios} 
\noindent
As a starting point, the user can either create a new or modify an 
existing scenario stored either locally or in a dedicated GitHub repository. 
CARJAN offers an initial set of key map templates for defining the grid-based layout 
of a traffic scenario in CARLA, such as a straight road segment, T-junction 
and intersection with no-go areas, sidewalks, two-lane roads, and decision 
boxes for entities. Entities are positioned in the grid via drag-and-drop.
Paths of defined agents in the scene are defined via waypoints in grid cells 
that are stepwise connected via cubic Bezier curves, from which they can deviate 
depending on their individual BT definitions in AJAN.
CARJAN offers specialized configuration options specific to the selected entity type.
Decision boxes for agents are defined by click-and-drag selection of a rectangular 
area on the grid and serve as interactive zones that emit signals upon collision, 
allowing the respective agent to react dynamically. 

\subsubsection{Modeling of intelligent agents and behaviors} 
\noindent
Each entity that is defined by the user to act in a CARLA traffic scenario, 
like a pedestrian, cyclist or car agent, gets automatically assigned to and 
initialized as an intelligent agent in AJAN. 
The behavior of such an agent is modeled by the user with an appropriate 
SPARQL Behavior Tree together with semantic knowledge of the agent
on the environment such as initial speed and contextual metadata encoded in a 
RDF triple store. For behavior definition, CARJAN offers the user a predefined 
set of actions within a BT. 
At runtime, the agents are linked to those behaviors and directives such as paths, 
hence the same behavior may cause different outcomes depending on the context.

\noindent
The agent knowledge is dynamically updated during simulation and processed within 
its BTs, which determine the event-driven, asynchronous action execution and decision-making 
of an agent in a CARLA simulated scene. For example, a pedestrian agent may
dynamically decide on when and which path to follow based on her observation
of the approaching car entering some defined regions on the road (decision boxes), 
or other sensed changes of the simulated scene.
In this respect, a pedestrian agent BT might allow the following event-driven action sequence: 
the pedestrian first walks to the initial critical waypoint of her path at her decision box, 
initiates a shoulder-check animation, adjusts her body posture (e.g., leaning forward to indicate 
crossing intention), and then decides on which path to follow further based on a SPARQL query 
considering the distance to the vehicle, the sensed car behavior such as slowing down or 
changed wheel stance, and the user-defined behavior strategy for the pedestrian.

\noindent
CARJAN also offers a set of parameterized high-level services that can be 
used for coding behaviors in the agent BTs including body pose 
animations. External AI action planning or learning methods can be easily 
integrated into agent SPARQL Behavior Trees via the plug-in concept of AJAN. 

\subsubsection{Data storage and management}

\noindent
On the CARJAN server side, persistent data of scenario descriptions are
stored in a dedicated scenario repository as RDF quads in TriG serialization format, 
which complements the RDF-encoded individual agents' knowledge and data storage 
in the respective AJAN triple stores.
This structured representation captures all scenario elements and their attributes and is 
maintained in a dedicated named graph. The collection of named graphs forms a local 
repository. Both individual scenarios and the entire collection of scenarios can be exported 
to the local machine or shared via GitHub. 

\noindent
For temporal scenario data changes by the user in the interactive JavaScript-based 
front-end of CARJAN, data manipulation can be done in JSON separate to the persistent 
triple store. The application itself relies on a state mechanism that 
provides short-term storage areas for user-generated or dynamically loaded content. 
Whenever modifications occur, they are captured in the state, and the resulting changes can
be written back to quads when a user opts to save or export data.
This approach helps to reduce performance overhead by avoiding frequent queries to the scenario 
repository and enables immediate visual feedback, for example when paths, symbols, or colors 
are updated on the grid surface in the CARJAN GUI. The interplay between the CARJAN server’s 
persistent store and the browser’s transitory state ensures that data remains consistent 
across all levels. As mentioned above, CARJAN also offers GitHub-based workflows for sharing 
scenario descriptions and allowing developers to collaborate or work on them in separate 
branches before merging in final versions.

\subsubsection{Automated translation to CARLA} 
\noindent
While the user of CARJAN in the generation process focuses on the modeling of 
traffic scenarios and agent behaviors, their translation into CARLA-compatible formats 
and configuration for testing (simulation) or storage is fully automatic, 
a one-click operation for the user in the CARJAN GUI. The internal CARJAN service 
{\tt carjanService} responsible for the translation is implemented using Flask.
In fact, {\tt carjanService} interfaces between the visual scenario editor of CARJAN, 
the AJAN agent execution backend and the CARLA simulation environment via dedicated 
Flask services. It handles all necessary conversions and coordinates re-initialization tasks, 
ensuring that updated scene configurations are immediately available for simulation. 

\noindent
If the user signals CARJAN to generate the current status of a modeled scenario for its 
simulation in CARLA, the {\tt carjanService} initializes a Flask instance to establish an 
interface to the CARLA environment. In a first step, the service waits for Flask to find the 
executable CARLA file, whose path is stored in an additional environment file for Node, along 
with other local environment variables such as the last scenario called in the frontend or 
GitHub login data. Flask then starts CARLA with an extended timeout to ensure that the simulator 
is fully booted and has established a stable client connection before continuing. After 
successful startup, Flask notifies {\tt carjanService} via an HTTP response, which in turn prompts 
the frontend to synchronize and returns the previously exported quads in their JSON-translated form 
in exchange. 

\noindent
Once both Flask and CARLA are ready, the generation sequence unfolds via the routes 
provided by Flask, which ultimately enables the integrated simulation of the scenario in CARLA
within CARJAN. This transforms the translation step, which is often tedious and repetitive for 
scene developers in traditional workflows, into a single-click process that enables iterative 
experimentation without the need for multiple rounds of Python scripting or reloading the environment.

\subsection{Integrated Scenario Simulation}
\noindent
After and at each time during the interactive generation of a scenario in CARJAN, 
the user can simulate the current status of the modeled scenario in CARLA and 
watch it live in the CARJAN GUI, together with a simultaneously displayed 
behavior tree execution of a selected agent in the scenario (cf. Fig. 3). 

\begin{figure}[ht]
\centering
\includegraphics[width=\textwidth]{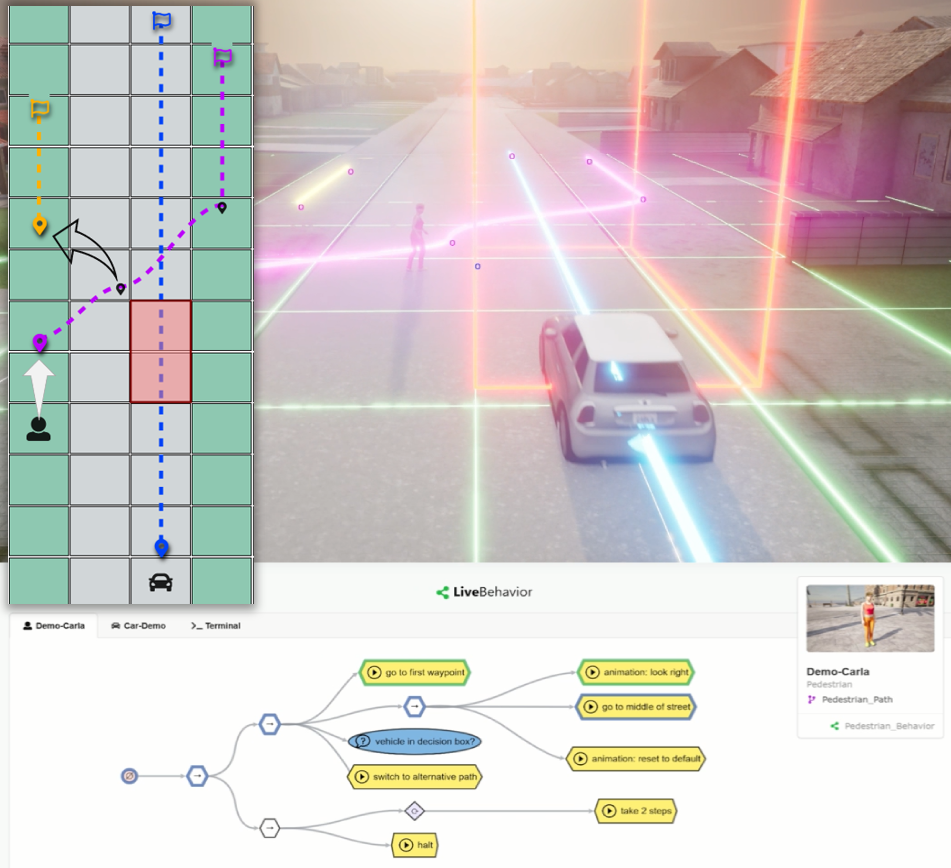}
\caption{Integrated simulation of scenario during its generation 
 together with behavior tree execution monitoring of selected pedestrian agent in the CARJAN GUI.} 
\label{fig3}
\end{figure}


\noindent
In particular, the integrated tool {\tt LiveBehavior} facilitates the explainability of behavioral 
decisions by enabling real-time visualization of behavior trees during execution: Each node in the 
behavior trees of each agent displays its current status—inactive, executed, successful, or 
failed—in an intuitive, color-coded interface. In addition, a runtime console displays relevant 
execution logs and agent status information, helping users correlate tree updates with observed 
agent behavior in the simulation.

\noindent
The execution of behaviors of multiple agents in a scenario is based on two main 
communication paradigms with CARLA via Flask services: synchronous and asynchronous actions. 
Synchronous actions confirm execution immediately and are ideal for instant state changes such as 
quick adjustments to speed or orientation. Asynchronous actions support extensive tasks such as 
piecewise navigation of paths per control point or the execution of detailed animations. For the 
latter, Flask uses dedicated threads linked to callback URIs so that actions can continue 
independently until completion or cancellation.

%
\section{Related Work}

\noindent
There are relevant approaches for traffic scenario generation and 
simulation for CARLA, in particular SCENIC and OASIS. 
While OASIS \cite{Carla-OASIS,OASIS25} is an impressive 
framework for visual authoring, deployment and diagnosis of traffic scenarios 
for CARLA, unlike CARJAN, it lacks support of a declarative, more explainable 
(BT-based glass-box) modeling and testing of behaviors and interactions 
of intelligent agents in these scenarios. 
SCENIC \cite{SCENIC19,SCENIC23} provides a powerful scripting language for 
probabilistic scene generation for CARLA based on a given set of user-defined scripts 
including the need for the developer to handle dependencies, but features neither a 
graphical user interface, nor integrated generation and simulation, nor 
declarative agent behavior modeling for scenarios.  

\section{Conclusion}

\noindent
We presented the novel, open-source available tool CARJAN for integrated agent-based
generation and simulation of traffic scenarios in CARLA.
It allows developers to transparently model and simulate scenarios within one graphical 
user interface, as well as to test different (hybrid) learning and planning methods 
for different agent behaviors. Current work on its second version includes improved 
support of benchmark maintenance, more scene layout and agent templates.

\begin{credits}
\subsubsection{\ackname} 
This work has been funded by the German Ministry for Research, Technology 
and Space (BMFTR) in the project MOMENTUM (01IW22001), 
and the European Commision in the project InnovAIte (09I02-03-V01-00029).

\subsubsection{\discintname}
The authors have no competing interests to declare that are
relevant to the content of this article.
\end{credits}

%
%
%
%

\end{document}